# APPLYING R-SPATIOGRAM IN OBJECT TRACKING FOR OCCLUSION HANDLING

Niloufar Salehi Dastjerdi and M. Omair Ahmad

Department of Electrical and Computer Engineering,
Concordia University, Montreal, Quebec, Canada
*ni_saleh@encs.concordia.ca*
*omair@ece.concordia.ca*

## ABSTRACT

*Object tracking is one of the most important problems in computer vision. The aim of video tracking is to extract the trajectories of a target or object of interest, i.e. accurately locate a moving target in a video sequence and discriminate target from non-targets in the feature space of the sequence. So, feature descriptors can have significant effects on such discrimination. In this paper, we use the basic idea of many trackers which consists of three main components of the reference model, i.e., object modeling, object detection and localization, and model updating. However, there are major improvements in our system. Our forth component, occlusion handling, utilizes the r-spatiogram to detect the best target candidate. While spatiogram contains some moments upon the coordinates of the pixels, r-spatiogram computes region-based compactness on the distribution of the given feature in the image that captures richer features to represent the objects. The proposed research develops an efficient and robust way to keep tracking the object throughout video sequences in the presence of significant appearance variations and severe occlusions. The proposed method is evaluated on the Princeton RGBD tracking dataset considering sequences with different challenges and the obtained results demonstrate the effectiveness of the proposed method.*

## 1. INTRODUCTION

Appearance feature description plays a crucial role in visual tracking, as the quality of the description directly affects the quality of the tracking performance. In general, the most desirable property of a feature description is to make the object easily distinguishable against non-targets in the feature space. From one pixel within a color image, the RGB color features can be naturally extracted. It is then not difficult to transform them into other color spaces or to gray levels. In addition, gradient and text features can also be extracted by considering the pixel within a local neighborhood. In order to describe a region of pixels in a higher level, one popular way is to use a descriptor based on statistics, such as the histogram which has been widely used in many computer vision applications to represent the pixel feature distribution. The histogram descriptor is a nonparametric estimation of the distribution over pixel values in a region. It owns a simple form and shows good robustness against translation and rotation. In [1], a generalized histogram called spatiogram was proposed to capture not only the values of the pixels but also their spatial relationships as well. To calculate the histogram efficiently, [2, 3] proposed intermediate representations to extract histograms called the integral histogram and the distributive histogram respectively. Although the histogram can accommodate any feature one at a time, the joint representation of several different features through histogram results in an exponential load as the number of features increases. In order to overcome the drawback of spatiogram, Dastjerdi et al. [4] introduced r-spatiogram based on the region-based compactness on the distribution of the given feature. In addition, a modification to the similarity measure for a spatiogram is respectively proposed by the authors. Choe et al. [5] introduced geogram that contains information about the perimeter of grouped regions in addition to features in the spatiogram. The author tested their feature descriptor and measured in object tracking scenario. . In [6] 13 different linear combinations of R, G, B pixel values were used to approximate 3D RGB color space using a set of 1D histograms is cheaper. Incremental Discriminative Color Tracking (IDCT) tracker [7] is another RGB tracker which utilized a discriminative method to provide an incremental object color modeling to separate the object from surrounding background in each frame. In [8], a fixed number of object parts are dynamically updated to account for appearance and shape changes. Ning et al. [9] proposed a mean-shift tracker with Corrected Background Weighted Histogram (CBWH). CBWH reduced background's interference in object localization by computing a color histogram with new weights to pixels in the target candidate region. He et al. [10] presented Locality Sensitive Histogram Tracker (LSHT) which computes a locality sensitive histogram that is computed at each pixel location along with a floating-point value corresponds with each bin to save the occurrence of an intensity value. Dastjerdi et al. [11], proposed an RGB-D tracker. The author utilized a combination of several features and applied a depth spatiogram to localize the object throughout the sequence. In this paper, we aim to extend our previous study [11] by applying r-spatiogram descriptor to improve the performance of the tracker.

The rest of the paper is organized as follows. Section 2 provides the preliminary theories of the r-spatiogram descriptor. Section 3 describes the details of the proposed method. Experimental results and performance evaluations are provided in Section 4. Finally, the conclusion are drawn in Section 5.

## 2. PRELIMINARY THEORIES

Feature description plays a crucial role in visual tracking, as the quality of the description directly affects the quality of the tracking performance. In the following subsection, preliminary theories about a newly feature descriptor, r-spatiogram, is described which we aim to use it in our tracking model.

### 2.1. r-spatiogram

Spatiogram is often calculated on the whole image region, which lacks robustness as a global image region is not discriminative enough to describe a shape. Hence, given an image that is a two-dimensional mapping $I: P \to v$ from pixels $P = [x, y]$ to values $v$, the third order r-spatiogram of an image, $rs$, is represented as,

$$rs_b^3 = \langle n_b, \mu_b, \Sigma_{b,} r_b \rangle, \qquad b = 1, \dots, B \qquad (1)$$

where $n_b$ shows the number of pixels whose value is that of the $b$th bin, $B$ denotes the number of bins in the spatiogram, $\mu_b$ and $\Sigma_b$ are the mean vectors and the covariance matrices of the coordinates of those pixels, respectively. $r$ is computed as (2) and denotes the ratio of the each subregion's histogram, $r_i n_b$, to whole image's histogram, $In_b$.

$$\boldsymbol{r_{ib}} = \frac{r_i n_b}{In_b}, \qquad b = 1, \dots, B; i = 1, \dots, M \qquad (2)$$

where M is the number of subregions in the image. In order to obtain the similarity between two r-spatiogram, the final similarity measure, $\rho$, is computed as (3),

$$\rho = \sum_{b=1}^{B} s \sqrt{n_b n'_b} \left[ 8\pi |\Sigma_b \Sigma'_b|^{1/4} N(\mu_b; \mu'_b, 2(\Sigma_b + \Sigma'_b)) \right] \tag{3}$$

$$rDist = \|r_b - r'_b\|_1 = \sum |r_{ib} - r_{ib'}|,$$
$$s = |1 - rDist| \tag{4}$$

Where $s$ denotes the similarity between ratios of two image, $r'_b$ and $r'_b$ and is computed with utilizing the l1-norm according to (4).

## 3. PROPOSED METHOD

We extend our previous study [11] to provide more accurate system. For this, we use the basic idea of the tracking model which consists of object modelling, object detection and localization, and model updating. Note that, in the occlusion state, different from [11], we apply r-spatiogram for better handling the occlusion. Using image features of object region and background region surrounding the object, an object model is created similar to [7, 11], to represent the target. Positive part of the log-likelihood ratio of the object and background determines the object model as in (5),

$$LR_{fb} = max\{ln \frac{max\{H_{obj}(p), \varepsilon\}}{max\{H_{bg}(p), \varepsilon\}}, 0\} \tag{5}$$

$$p = \alpha * fb$$

where $H_{obj}(p)$ is the histogram computed within the object rectangle, and $H_{bg}(p)$ is the histogram of background rectangle which is the region surrounding the object. Index $fb$ indicates the feature bin and ranges from 1 to total number of histogram bins. $\alpha$ is the percentage of randomly selected bins and we apply it to improve the speed. $\varepsilon$ is a small nonzero value to avoid dividing by zero. Different from [11] which authors used combination of several features to create feature vector, in this paper R (red), G (green), and B (blue) channels of image are utilized to create the feature vector as in (6),

$$\mathbf{F}(x, y) = [R, G, B] \tag{6}$$

In parallel, the depth image is pre-processed to normalize the depth values between 0 and 255. The closer to the camera, the larger the depth value is. In addition, similar to [12] which introduced a fast tracker, a depth segmentation approach is done by applying K-means to estimate initial clusters or regions of interest (ROI), followed by, connected component analysis that is analyzed in the image plane to distinguish between objects located within the same depth plane. So similar to [11] a connected component matrix ($CCR$) is obtained to use in the object localization. In the next step, the positive log-likelihood ratio LR, is computed as a mapping function to provide an intermediate map ($IM$), from the object region,

$$IM(x_i, y_j) = LR(\mathbf{F}(x_i, y_j)) \tag{7}$$

where $(x_i, y_j)$ is pixel coordinate and $\mathbf{F}(x_i, y_j)$ shows feature vector, in the object region. Then, employing the depth data, we create a masked similar to [11]. It is computed according to (8) and alleviates the problem of same color or texture in background which may lead to drift in tracking.

$$M(x_i, y_j) = IM(x_i, y_j) * CCR(x_k, y_m) \tag{8}$$

$CCR(x_k, y_m)$ denotes the matrix of connected component with the same size of $IM$ and includes zero values for coordinates containing non-targets. So, $(x_k, y_m)$ indicates pixel coordinate which is non-zero and has same depth as target's depth.

In order to localize the object, mean-shift algorithm is applied on masked map which obtained from (8). The centroid of the masked map of the object in the current frame is used to localize the object in the next frame. At each iteration, the center of the object rectangle is shifted to the center of the current masked map of the object. The iteration will repeated until the object is placed inside the rectangle completely. At this step, we apply the r-spatiogram descriptor on the BB to provide the accurate representation of the tracked object as (9).

$$RS_b^3 = \langle n_b, \mu_b, \Sigma_b, r_b \rangle, \qquad b = 1, \dots, B \tag{9}$$

Note that, in this study we do not compute the spatial covariance of the r-spatiogram descriptor to improve the speed and make a right balance between complexity and speed. In order to update the target model to the recent observations, positive log-likelihood ratio at current frame ($LR^t$) is used similar to [7, 11]. Once the object location at the current frame is computed by the mean-shift, $LR^t$ is applied to update the previous object model, $LR^{t-1}$. So, the updated object model, $LR^{t+1}$, is computed as (10),

$$LR^{t+1} = \rho \times LR^t + (1-\rho) \times LR^{t-1} \tag{10}$$

$\rho$ is a forgetting factor to make the balance between the old and new observations and is set to 0.1. To recover from the occlusion state at each occluded frame, depth spatiogram of the region inside bounding box is analysed similar to [11]. For this, in order to locate the occluder's position in the current bounding box, we compute the depth spatiogram to obtain the depth value and depth location of the occluder. It should be noted that the use of the image coordinates with real-world depth facilitates defining the search space and yields a more accurate estimate of the position of the occluded target.

Suppose the target is detected near-fully occluded in the current frame $t$. First, a search region is defined at the centroid of occluder. Then, the area around the occluder is searched to locate the target in the next frame $t + 1$. The object has a newly rising peak with a depth value smaller than the depth value of the occluder. So according to the spatial information obtained from depth spatiogram, the object is highly likely to re-emerge in those image areas close to the center of obtained spatial information. We then create a new BB centered at the obtained spatial means, and covering the obtained depth around the center, as object candidate. So the size of the new BB is correspondence to the number of pixels whose value is that of the obtained depth. For the object candidate, the r-spatiogram descriptor is created according to the explanation in section 2.1 and its similarity, $S$, is compared with object in the previous frame using r-spatiogram similarity measure which described in equation (3). If the similarity of the candidate is greater than a given threshold (95% in this study), it is regarded as object. Otherwise, we expand the size of candidate's BB to create a search area, and search for a region having most similar features as target. The search is performed by sliding-window from left to right and from top to bottom in the search area. The window size is fixed and is equal to the size of candidate's BB before expanding. The search window jumped horizontally 10% of the width or vertically 10% of the height of the search area. Then, the best 10% matchings are extracted as our candidates, and the dissimilarity of the object and all candidates are computed. Finally, the region with the smallest distance is selected as the best candidate. This candidate is passed to the second step, i.e., object detection and localization, and the algorithm will be continued to track the object throughout the sequence.

## 4. EXPERIMENTAL RESULTS

We evaluate our method on Princeton Tracking RGBD dataset [13] which consists of test sequences with different kind of challenges such as cluttered background, partial and complete occlusion, fast movement, shape deformation and distortion, and so on. In order to provide a fair and consistent comparison to what has been done in [11], we compare the performance of the proposed method with the tracking methods, including CBWH [9], LSHT [10], IDCT tracker [7], and our old tracker [11].

### 4.1 Objective Results

To evaluate the performance of our proposed method, two objective measures are used. Average center location error and the average overlap rate. The average center location error (ACLE) [14] is a widely used metric that computes the average Euclidean distance according to (13),

$$ACLE = \frac{1}{n}\sum_{i=1}^{n}\sqrt{(X_i - X_i^g)^2 + (Y_i - Y_i^g)^2} \qquad (11)$$

where $[X_i, Y_i]$ denotes the center location of the object obtained by the tracker which is determined by the central point of the object rectangle. $[X_i, Y_i]$ indicates the center of the ground truth rectangle. $n$ is the total number of frames and $i$ ranges from 1 to $n$. The average overlap ratio (AOR) measures the overlap ratio between the estimated BB predicted from the tracker ($B_t$) and the annotated BB ($B_t^g$) according to (12),

$$AOR = \frac{B_t \cap B_t^g}{B_t \cup B_t^g} \qquad (12)$$

The results evaluated by the above measures are shown in Table 1. As our proposed method utilized r-spatiogram descriptor to boost the performance of tracker mostly in sequences with occlusions, we provide the comparison results for those sequences. As can be seen, our method achieved the best scores in all sequences except "zcup_move_1" and "hand_occ" which obtained the second best performance. In "zcup_move_1", the main challenge is similar background and due to use of less features than what we used in our old tracker, our proposed method achieves lower accuracy which is clear in Figure 2. Also, in "hand_occ", one hand (object) is occluded by the other hand. Therefore, considering that we applied a small number of features, the performance is better using our old tracker which used a combination of several features.

Table 1. The average center location errors (ACLE) and average overlap rate (AOR) of the evaluated methods on the sequences.

| Algorithm | Seq. | bear_front | child_no1 | face_occ5 | New_ex_occ4 | zcup_move_1 | dog_occ_2 | express1_occ | library2.1_occ | hand_occ |
|---|---|---|---|---|---|---|---|---|---|---|
| CBWH | ACLE | 20.4 | 31.6 | 18.2 | 25.6 | 18.4 | 21.6 | 32.1 | 17.7 | 19.3 |
|  | AOR | 0.59 | 0.52 | 0.60 | 0.75 | 0.68 | 0.61 | 0.4 | 0.65 | 0.75 |
| LSHT | ACLE | 26.1 | 35.0 | 21.4 | 26.0 | 17.1 | 23.5 | 27.3 | 20.4 | 19 |
|  | AOR | 0.75 | 0.49 | 0.61 | 0.78 | 0.81 | 0.5 | 0.47 | 0.52 | 0.77 |
| IDCT | ACLE | 12.3 | 16.9 | 25.0 | 27.2 | **6.7** | 18.2 | 25 | 22 | 17.8 |
|  | AOR | 0.67 | 0.72 | 0.78 | 0.63 | **0.89** | 0.72 | 0.52 | 0.45 | 0.8 |
| Our old tracker | ACLE | 4.5 | 9.9 | 10.8 | 19.7 | 16.4 | 11.8 | 14.2 | 12.6 | **15.5** |
|  | AOR | 0.82 | 0.66 | 0.92 | 0.84 | 0.52 | 0.87 | 0.7 | 0.81 | **0.88** |
| proposed method | ACLE | **3.8** | **8.7** | **6.3** | **11.2** | 17.4 | **8.7** | **13.8** | **10.8** | 16.8 |
|  | AOR | **0.92** | **0.8** | **0.96** | **0.93** | 0.45 | **0.91** | **0.77** | **0.88** | 0.82 |

The details of center location errors for two sequences is shown in Figure 1. Note that frames with occlusion are considered to be indicated in the figure to show the effectiveness of the proposed model.

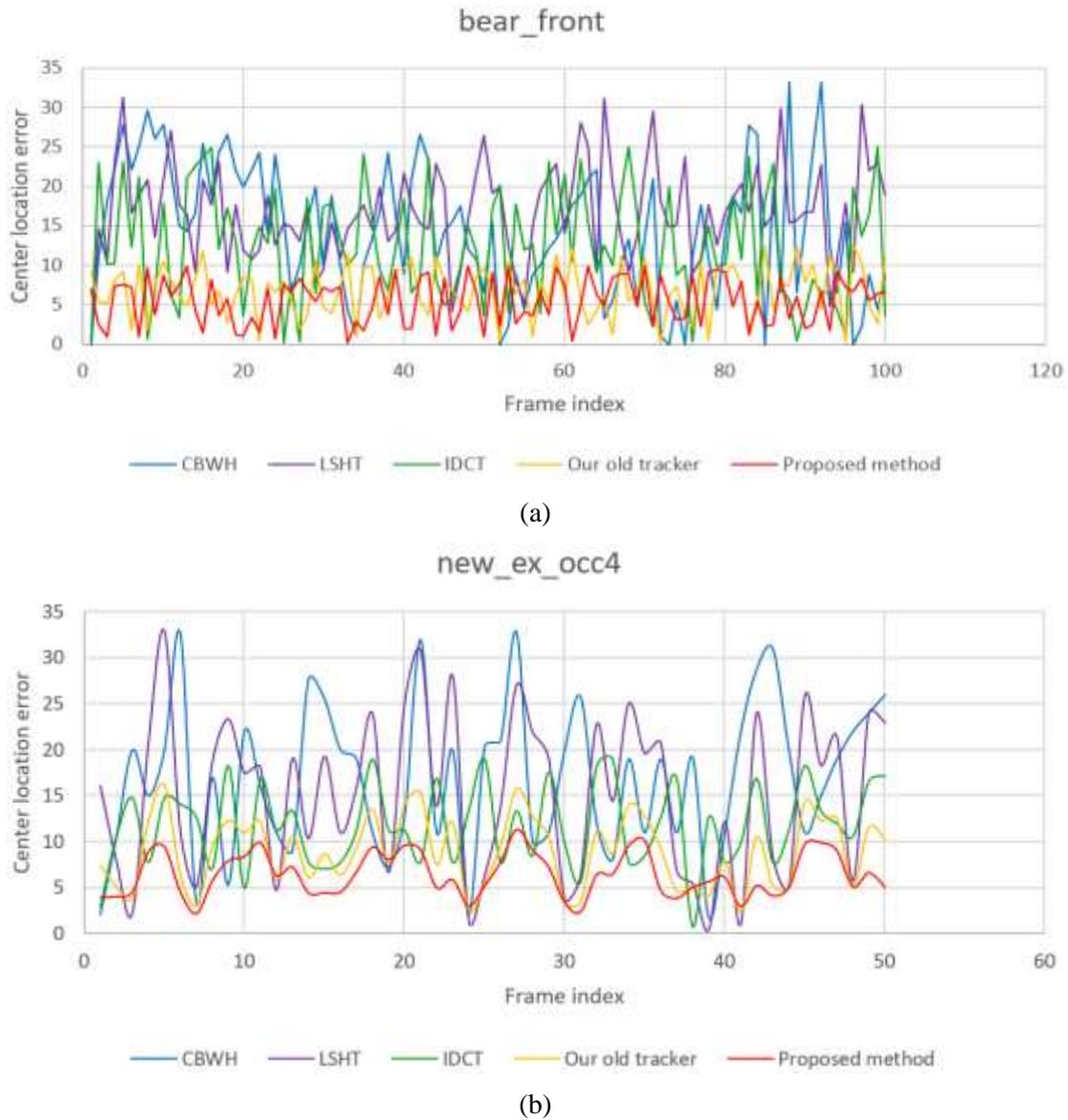

Figure 1. The center location errors of the sequences (a) "bear_front" and (b) "New_ex_occ4".

## 4.2 Subjective Results

Figure 2 shows sample screenshots of the tracking results for the sequences "bear_front", "New_ex_occ4", and "zcup_move_1" in top, middle, and bottom rows, respectively. Red rectangle shows the tracked object using our proposed method. The yellow rectangle indicates the tracked object applying our old tracker. Results demonstrate that our proposed method effectively tracks the object throughout the sequence.

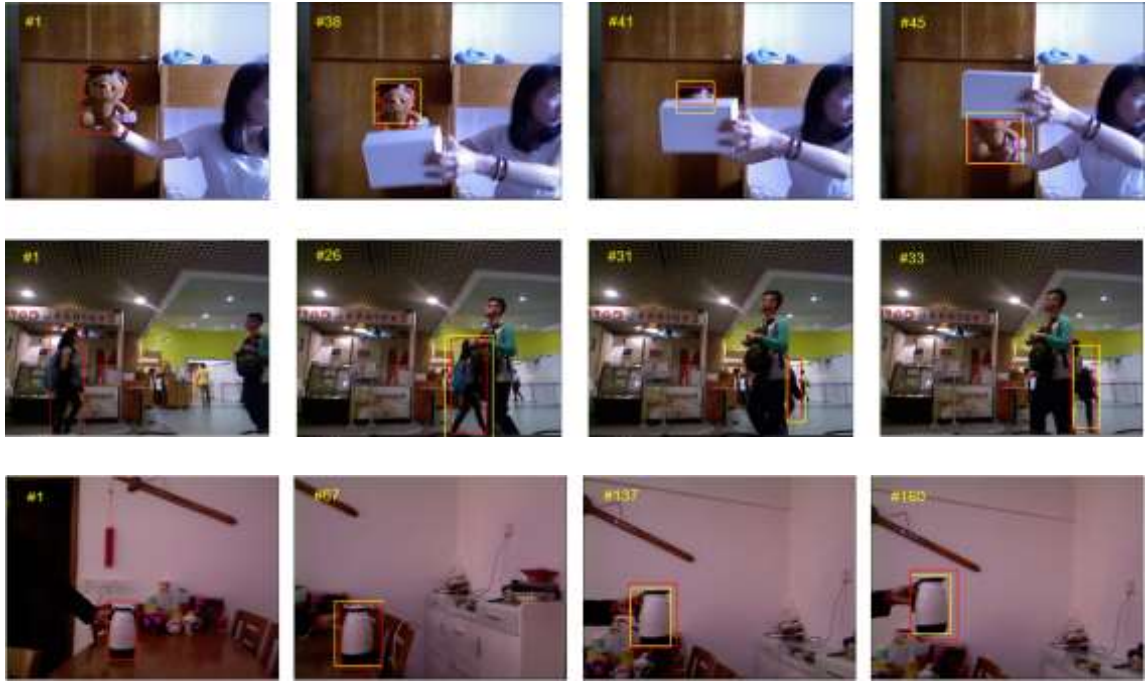

Figure 2. Sample screenshots of the tracking results using our old tracker (yellow rectangle) and the proposed method (red rectangle).

## 5. CONCLUSIONS

A tracking algorithm infers the location of a target from a sequence of measurements. In tracking method, the main challenge lies in the design of measurements, i.e., features, effective at tracking performance. So, role of features in representing the appearance of the object is one of the key building blocks in tracking. In this paper, we used the minimum components of common tracking frameworks and few number of features to make a right balance between complexity and accuracy. In order to handle the problem of occlusion, r-spatiogram descriptor was utilized. Experimental results demonstrated the effectiveness of the proposed method especially in sequences with occlusions. Future work should be aimed at efficient combination of features. Different feature combinations generally produce different detection performances and previous works generally reported better results using more features. So, how to select proper feature set for detecting a specific object to ensure good performance in terms of detection accuracy can be considered in further work. In addition, combining depth with r-spatiogram may improve the accuracy of tracker in sequences with complex challenges.

**Authors**

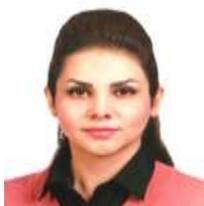

**Niloufar Salehi Dastjerdi** received the M.Sc. degree in computer science in 2012 from University Technology Malaysia. She is currently working toward the Ph.D. degree in the department of Electrical and Computer Engineering, Concordia University, Montreal, Canada. Her research interests include visual tracking, computer vision, image processing, and pattern recognition.

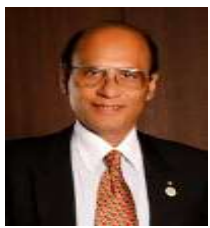

**M. Omair Ahmad** (S'69 M'78–SM'83–F'01) received the B.Eng. degree from Sir George Williams University, Montreal, QC, Canada, and the Ph.D. degree from Concordia University, Montreal, QC, Canada, both in electrical engineering. From 1978 to 1979, he was a Faculty Member with the New York University College, Buffalo, NY, USA. In September 1979, he joined the Faculty of Concordia University as an Assistant Professor of computer science. He joined the Department of Electrical and Computer Engineering, Concordia University, where he was the Chair with the department from June 2002 to May 2005 and is currently a Professor. He holds the Concordia University Research Chair (Tier I) in Multimedia Signal Processing. He has authored or coauthored extensively in the area of signal processing and holds four patents. His current research interests include the areas of multidimensional filter design, speech, image and video processing, nonlinear signal processing, communication DSP, artificial neural networks, and VLSI circuits for signal processing. He was a Founding Researcher at Micronet, Ottawa, from its inception in 1990 as a Canadian Network of Centers of Excellence until its expiration in 2004. Previously, he was an Examiner of the order of Engineers of Quebec. He was an Associate Editor for the IEEE TRANSACTIONS ON CIRCUITS AND SYSTEMS PART I: FUNDAMENTAL THEORY AND APPLICATIONS from June 1999 to December 2001. He was the Local Arrangements Chairman of the 1984 IEEE International Symposium on Circuits and Systems. In 1988, he was a member of the Admission and Advancement Committee of the IEEE. He has served as the Program Co-Chair for the 1995 IEEE International Conference on Neural Networks and Signal Processing, the 2003 IEEE International Conference on Neural Networks and Signal Processing, and the 2004 IEEE International Midwest Symposium on Circuits and Systems. He was a General Co-Chair for the 2008 IEEE International Conference on Neural Networks and Signal Processing. He is the Chair of the Montreal Chapter IEEE Circuits and Systems Society. He is a recipient of numerous honors and awards, including the Wighton Fellowship from the Sandford Fleming Foundation, an induction to Provosts Circle of Distinction for Career Achievements, and the Award of Excellence in Doctoral Supervision from the Faculty of Engineering and Computer Science of Concordia University.